\documentclass[11pt,letterpaper]{article}

\usepackage{authblk}
\usepackage{conll2018}
\usepackage{pslatex}
\usepackage[english]{babel}
\usepackage[utf8]{inputenc}
\usepackage{amsmath}
\usepackage{bm}
\usepackage{graphicx}
\usepackage{tikz}
\usepackage{xcolor}
\usepackage{url}
\usepackage{rotating}
\usepackage{natbib}
\usepackage{amssymb}
\usepackage{linguex}

\newcommand{\key}[1]{\textbf{#1}}
\newcommand{\soft}[1]{}
\newcommand{\nopreview}[1]{}

\newcommand{\gulordavarnn}{GRNN\xspace}
\newcommand{\googlernn}{JRNN\xspace}

\aclfinalcopy

\title{Do RNNs learn human-like abstract word order preferences?}

\author[1]{\textbf{Richard Futrell}}
\author[2]{\textbf{Roger P. Levy}}

\affil[1]{Department of Language Science, UC Irvine, \tt{rfutrell@uci.edu}}
\affil[2]{Department of Brain and Cognitive Sciences, MIT, \tt{rplevy@mit.edu}}

\date{}

\begin{document}

\maketitle

\begin{abstract}

  RNN language models have achieved state-of-the-art results on various tasks, but what exactly they are representing about syntax is as yet unclear. Here we investigate whether RNN language models learn humanlike word order preferences in syntactic alternations. We collect language model surprisal scores for controlled sentence stimuli exhibiting major syntactic alternations in English: heavy NP shift, particle shift, the dative alternation, and the genitive alternation. We show that RNN language models reproduce human preferences in these alternations based on NP length, animacy, and definiteness. We collect human acceptability ratings for our stimuli, in the first acceptability judgment experiment directly manipulating the predictors of syntactic alternations. We show that the RNNs' performance is similar to the human acceptability ratings and is not matched by an $n$-gram baseline model. Our results show that RNNs learn the abstract features of weight, animacy, and definiteness which underlie soft constraints on syntactic alternations.
  
\end{abstract}

The best-performing models for many natural language processing tasks in recent years have been recurrent neural networks (RNNs) \citep{elman1990finding,sutskever2014sequence,goldberg2017neural}, but the black-box nature of these models makes it hard to know exactly what generalizations they have learned about their linguistic input: Have they learned generalizations stated over hierarchical structures, or only dependencies among relatively local groups of words \citep{linzen2016assessing,gulordava2018colorless,futrell2018rnns}? Do they represent structures analogous to syntactic dependency trees \citep{williams2018latent}, and can they represent complex relationships such as filler--gap dependencies \citep{chowdhury2018rnn,wilcox2018what}? In order to make progress with RNNs, it is crucial to determine what RNNs actually learn given currently standard practices; then we can design network architectures, objective functions, and training practices to build on strengths and alleviate weaknesses \citep{linzen2018what}.

In this work, we investigate whether RNNs trained on a language modeling objective learn certain syntactic preferences exhibited by humans, especially those involving word order. We draw on a rich literature from quantitative linguistics that has investigated these preferences in corpora and experiments \citep[e.g.,][]{mcdonald1993word,stallings1998phrasal,bresnan2007predicting,rosenbach2008animacy}. 

Word order preferences are a key aspect of human linguistic knowledge. In many cases, they can be captured using local co-occurrence statistics: for example, the preference for subject--verb--object word order in English can often be captured directly in short word strings, as in the dramatic preference for \emph{I ate apples} over \emph{I apples ate}. However, some word order preferences are more abstract and can only be stated in terms of higher-order linguistic units and abstract features. For example, humans exhibit a general preference for word orders in which words linked in syntactic dependencies are close to each other: such sentences are produced more frequently and comprehended more easily \citep{hawkins1994performance,futrell2015largescale,temperley2018minimizing}.

We are interested in whether RNNs learn abstract word order preferences as a way of probing their syntactic knowledge. If RNNs exhibit these preferences for appropriately controlled stimuli, then on some level they have learned the abstractions required to state them. 

Knowing whether RNNs show human-like word order preferences also bears on their suitability as language generation systems. \citet{white2012minimal} have shown that language generation systems produce better output when human-like word order preferecences are built in; it may turn out that RNN language models reproduce such preferences such that they do not need to be built in explicitly.

As part of this work, we validate and quantify these word order preferences for humans by collecting acceptability ratings for English sentences with different word orders. To our knowledge, this is the first experimental acceptability-judgment study of these word order preferences using fully controlled stimuli; previous experimental work has used naturalistic stimuli derived from corpora, in which the predictors of word order are not directly manipulated \citep{rosenbach2003aspects,bresnan2007syntactic}.

\subsection*{Alternations studied}

We study four syntactic alternations in English: \key{particle shift}, in which a verbal particle can appear directly after the verb or later (e.g., \emph{give up the habit} vs. \emph{give the habit up}); \key{heavy NP shift}, in which a verb is followed by an NP and a PP with order NP--PP or PP--NP; the \key{dative alternation} (e.g. \emph{give a book to Tom} vs. \emph{give Tom a book}); and the \key{genitive alternation} (e.g. \emph{the movie's title} vs. \emph{the title of the movie}). 

In all these alternations, three common factors influencing word order preferences are evident: short constituents go before long constituents; words which are definite go earlier; and words referring to animate entities go earlier. In fact these preferences are very general patterns across languages, and in some languages constitute hard constraints \citep{bresnan2001soft}.

\section{Methods}

We investigate the learned word order preferences of RNNs by studying the total probability they assign to sentences with various word order properties. Specifically, we create sentences by hand which can appear in a number of configurations, and study how these manipulations affect the \textsc{surprisal} value assigned by an RNN to a sentence. Surprisal is the negative log probability:
\begin{align*}
\nonumber
S(x_{i=1}^n) &= -\log_2 p(x_{i=1}^n) \\
&= - \sum_{i=1}^n \log_2 p(x_i|x_{j=1}^{i-1}),
\end{align*}
where $x_{i=1}^n$ is a sequence of $n$ words forming a sentence and the conditional probability $p(x_i|x_{j=1}^{i-1})$ is calculated as the RNN's normalized softmax activation for $x_i$ given its hidden state after consuming $x_{j=1}^{i-1}$.

Surprisal has a number of interpretations that make it convenient as a dependent variable for examining language model behavior. First, surprisal is equivalent to the contribution of a sentence to a language model's cross-entropy loss: effectively, our RNN language models are trained with the sole objective of minimizing the average surprisal of training sentences, so surprisal is directly related to the model's performance. Second, word-by-word surprisal has been found to be an effective predictor of human comprehension difficulty \citep{hale2001probabilistic,levy2008expectation,smith2013effect}; interpreting surprisal as metric of ``difficulty'' allows us to analyze RNN behavior analogously to human processing behavior \citep{vanschijndel2018modeling,futrell2018rnns}. Third, surprisal more generally reflects the dispreference or \key{markedness} of a sequence according to a language model. High surprisal for a sentence corresponds to a relative dispreference for that sentence. When the logarithm is taken to base 2, surprisal is equivalent to the bits of information required to encode a sentence under a model.

In the studies below, we test hypotheses statistically using maximal linear mixed-effects models \citep{baayen2008mixed,barr2013random} fit to predict surprisals given experimental conditions. 

\subsection{Models tested}

We study the behavior of two LSTMs trained on a language modeling objective over English text: the one presented in \citet{jozefowicz2016exploring} as ``BIG LSTM+CNN Inputs'', which we call ``\googlernn'', which was trained on the One Billion Word Benchmark \citep{chelba2013one} with two hidden layers of 8196 units and CNN character embeddings as input; and the one presented in \citet{gulordava2018colorless}, which we call ``\gulordavarnn'', with two hidden layers of 650 units, trained on 90 million tokens of English Wikipedia.

As a control, we also study surprisals assigned by an $n$-gram model trained on the One Billion Word Benchmark (a 5-gram model with modified Kneser-Ney interpolation, fit by KenLM with default parameters) \citep{heafield2013scalable}. The $n$-gram surprisals tell us to what extent the patterns under study can be learned purely from co-occurrence statistics with a small context window without any generalization over words. To the extent that LSTMs yield more humanlike performance than the $n$-gram model, this indicates one of two things. Either they have learned generalizations that are formulated in terms of more abstract linguistic features, or they have learned generalizations that can span larger distances than the $n$-gram window.

\subsection{Human acceptability ratings}

We also compare RNN surprisals against human preferences on our experimental items. We collected acceptability judgments on a scale of 1 (least acceptable) to 5 (most acceptable) over Amazon Mechanical Turk.\footnote{Preregistered at \url{https://aspredicted.org/sh9zf.pdf}.} For the studies of heavy NP shift, the dative alternation, and the genitive alternation, we collected data from 64 participants, filtering out participants who were not native English speakers or who could not correctly answer 80\% of simple comprehension questions about the experimental items. After filtering, we had data from 55 participants. For the study of particle shift, we used data from a previous (unpublished) acceptability rating experiment with 196 subjects, and the same filtering criteria. After filtering, we had data from 156 participants.

\section{Heavy NP Shift}

\textsc{Heavy NP shift} describes a scenario where constituent weight preferences become so strong that an order which would otherwise be unacceptable becomes more acceptable, as shown in Example~\ref{ex:heavy-np-shift}.

\ex. \label{ex:heavy-np-shift}
\a. The publisher announced a book on Thursday.
\b. *The publisher announced on Thursday a book.
\c. The publisher announced a new book from a famous author who always produced bestsellers on Thursday.
\d. The publisher announced on Thursday a new book from a famous author who always produced bestsellers.

In these examples, the verb \emph{announced} is followed by a noun phrase (\emph{a (new) book...}) and a temporal PP adjunct (\emph{on Thursday}). The usual order for these elements is to put the NP before the PP, but when the NP becomes very heavy, the PP might be placed closer to the verb, which case the word order is called \textsc{shifted}. Heavy NP shift is the primary example of locality effects in word order preferences, in that it creates shorter dependencies from the verb to the NP and the PP.

We tested whether RNNs show length-based preferences similar to Example~\ref{ex:heavy-np-shift}.\footnote{The preregistration for this experiment can be viewed at \url{https://aspredicted.org/ea6m8.pdf}.} We adapted 40 items from \citet{stallings1998phrasal} which consist of a verb followed by an NP and a temporal PP adjunct, where the order of the NP and the PP and the length of the NP are manipulated. If the networks show human-like word ordering preferences, there should be a penalty for PP--NP order when the NP is short, but this penalty should be smaller or nonexistent when the NP is long.

\begin{figure}
\includegraphics[scale=.45]{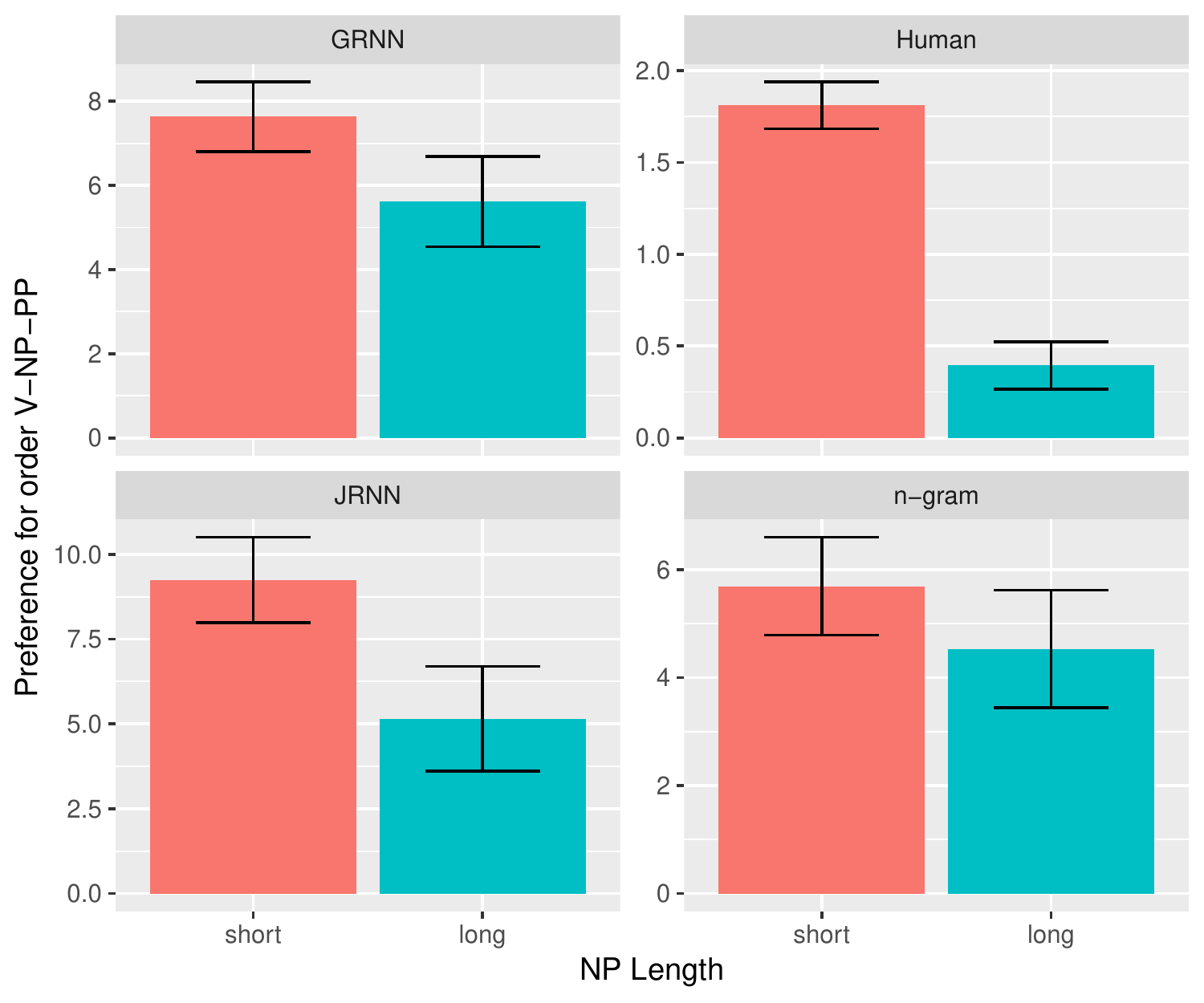}
\caption{Mean preference for standard word order by NP length. In this and other figures, for computational models, preference is measured as total sentence surprisal for Verb--NP--PP order minus total sentence surprisal for Verb--PP--NP order; error bars represent 95\% confidence intervals of the contrasts between conditions, computed by subtracting out the by-item means before calculating the intervals \citep{masson2003using}. For the human data, preference is measured as the difference in mean acceptability for Verb--PP--NP minus Verb--NP--PP, and error bars represent 95\% confidence intervals of the contrasts between conditions after subtracting out by-item and by-subject means.}
\label{fig:heavy-np-total}
\end{figure}

Figure~\ref{fig:heavy-np-total} shows the models' preference for the standard word order (Verb--NP--PP) over the shifted word order (Verb--PP--NP), calculated as the surprisal of sentences in shifted word order minus their surprisal in standard word order. Also included are the human acceptability ratings, where the preference for the order Verb--NP--PP is calculated as the average acceptability difference between Verb--NP--PP and Verb--PP--NP across items. In all cases, we see that the shifted order becomes more preferred when the NP is long, although it never becomes the most preferred order.

Our experimental design allows us to control for the effect of sentence length on RNN surprisals. The sentences with long NPs are naturally expected to have higher RNN surprisal than the ones with short noun phrases, since they have more words and thus higher information content. However, the Verb--NP--PP preference is quantified as the \emph{difference} between surprisal of order Verb--PP--NP and surprisal of the order Verb--NP--PP, for both the long and short NP conditions. The long NP conditions may have higher surprisal overall, but this will be cancelled out in the difference. The crucial question is then whether this preference is larger in the long NP case than in the short NP case---whether the red and blue values in Figure~\ref{fig:heavy-np-total} are significantly different across items. The crucial statistic for each item $i$ is given by the interaction $I_i$:
\begin{align*}
  \nonumber
  I_i = &(S_i(\text{\small{short}}, \text{\small{Verb--NP--PP}}) - S_i(\text{\small{short}}, \text{\small{Verb--PP--NP}})) \\
  &- (S_i(\text{\small{long}}, \text{\small{Verb--NP--PP}}) - S_i(\text{\small{long}}, \text{\small{Verb--PP--NP}})),
\end{align*}
where $S_i$ is the surprisal for the $i$th item in the given condition. If $I_i$ is significantly positive across items, then we have evidence that NP length causes a preference for Verb--PP--NP order even when controlling for the intrinsic effects of length and of the particular words in each item. The same logic is applied to the analysis of the acceptability ratings data. All studies in this paper apply this same design and analysis. Similar designs are common in psycholinguistics, and are applied to RNN surprisal data in \citet{futrell2018rnns} and \citet{wilcox2018what}. 

To test the significance of the interaction, we use mixed-effects modeling with random intercepts and slopes by item. We find that the interaction is statistically significant in \googlernn (interaction size $4.0$ bits, $p<0.001$) and \gulordavarnn ($2.0$ bits, $p = 0.02$), but not in the $n$-gram baseline ($1.2$ bits, $p = 0.13$). The interaction in \googlernn is significantly stronger than in the $n$-gram baseline ($p=0.03$), but the interaction in \gulordavarnn is not significantly stronger than the $n$-gram baseline ($p=0.48$). None of the models show the effect as strongly as the human acceptability judgments.

Thus we find that both \googlernn and \gulordavarnn exhibit human-like word order biases for Heavy NP shift, but do not find evidence for such a bias in the $n$-gram baseline. The result suggests that the LSTM models have learned a higher-order generalization that is not trivially present in $n$-gram statistics.

\section{Phrasal verbs and particle shift}
\label{sec:particle-shift}

Another domain of word order variation similar to Heavy NP shift is phrasal verbs, which consist of a verb and a particle, such as \emph{give up}. The object NP of a transitive phrasal verb can appear in two positions: it can be \textsc{shifted} (after the particle) or \textsc{unshifted} (before the particle). As in Heavy NP shift, the shifted order is generally preferred when the NP is long:

\ex. \label{ex:particle-shift}
\a. Kim gave up the habit. [shifted]
\b. Kim gave the habit up. [unshifted]
\c. Kim gave up the habit that was preventing success in the workplace. [shifted]
\d. Kim gave the habit that was preventing success in the workplace up. [unshifted]

The fact that both word orders are possible is called \textsc{particle shift}. Particle shift provides another arena to test whether RNNs have learned the basic short-before-long constituent ordering preference in English. Furthermore, particle shift is also affected by the animacy of the object NP, in that the unshifted order is preferred when the object NP is animate \citep{gries2003multifactorial}, so we can use this construction to test order preferences involving both length and animacy. 

We designed 32 experimental items consisting of sentences with phrasal verbs as in Example~\ref{ex:particle-shift}, where each item could occur with either a long or a short object NP. Long NPs were created by adding adjectives and postmodifiers to short NPs. Half of the items had inanimate objects; half had animate objects. All NPs were definite. We tested the effects of NP length, NP animacy, and word order on language model surprisal.\footnote{The preregistration for this experiment can be viewed at \url{https://aspredicted.org/uu7am.pdf}.} 

\begin{figure}
\centering
\includegraphics[scale=.45]{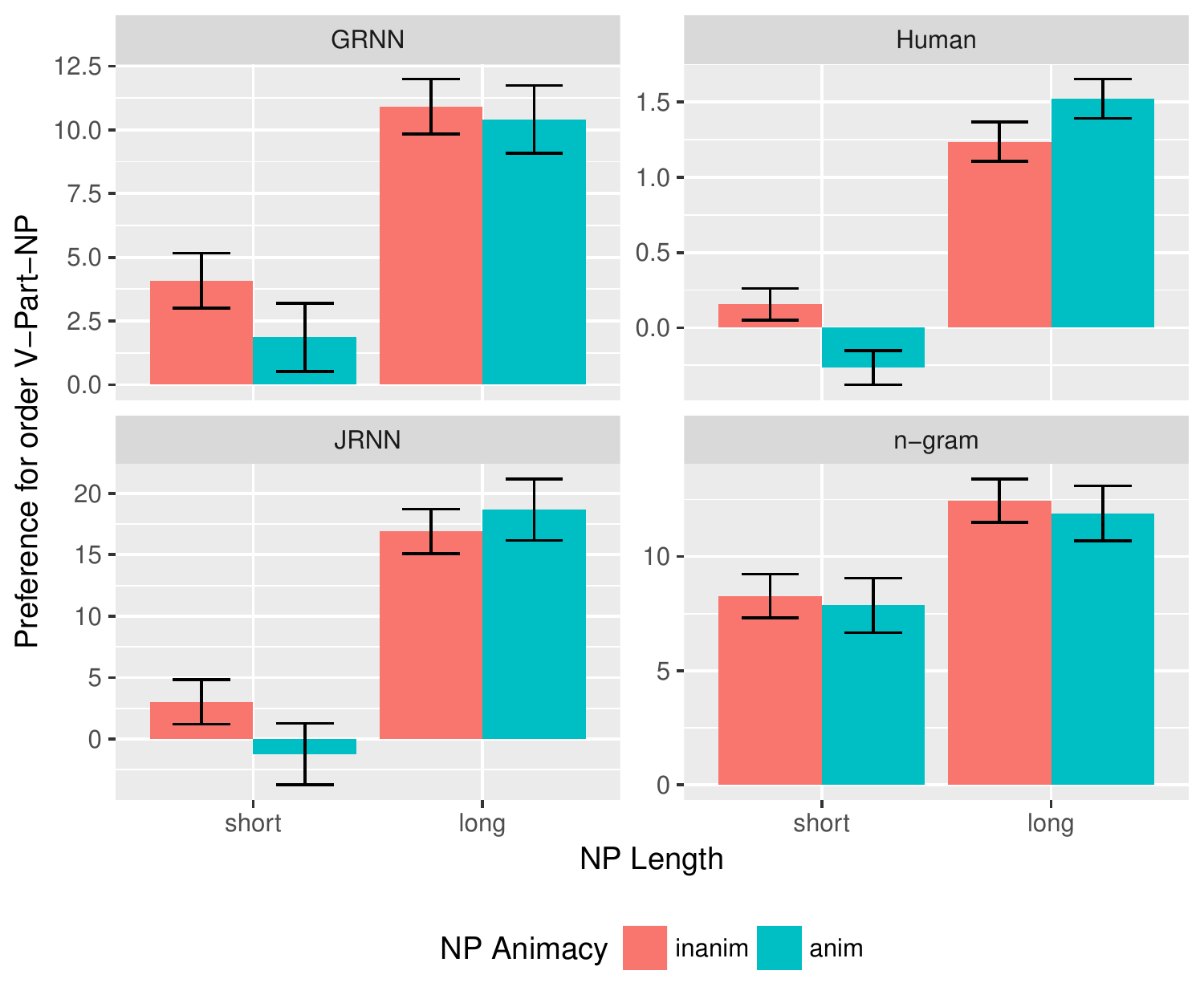}
\caption{Preference for shifted word order (total sentence surprisal for Verb--Particle--NP order minus Verb--NP--Particle order) by NP length, NP animacy, and model.}
\label{fig:particle-shift}
\end{figure}

Figure~\ref{fig:particle-shift} shows the average preference for shifted word order according to each model, calculated as the surprisal of the shifted order minus the surprisal of the unshifted order. In general, we see that when the object NP is long, the shifted order is relatively preferred; the effect is strongest in \googlernn. In regressions, we found that the interaction of NP length and word order is significant in \googlernn ($16.9$ bits, $p<0.001$), \gulordavarnn ($7.8$ bits, $p<0.001$), and the $n$-gram baseline ($4.1$ bits, $p<0.001$). However, the interaction in the $n$-gram baseline is significantly smaller than in \googlernn ($p<0.001$) and \gulordavarnn ($p<0.01$). 

The effects of animacy are unexpectedly intricate. Numerically, \gulordavarnn and the $n$-gram baseline show the expected effect: an animate NP favors unshifted order. However, in the human ratings we find that the expected animacy effect for short NPs actually reverses for long NPs, and this is reflected numerically in \googlernn. No effects of animacy are significant in any model. In the human data, animacy has a significant interaction favoring the unshifted word order for short NPs ($p<0.001$) with an interaction that reverses the effect for long NPs ($p<0.001$). This reversal is surprising given what was previously known about word order in phrasal verbs.

Our investigation of particle shift has shown that LSTM models learn short-before-long length preferences in regard to word order in phrasal verb constructions; $n$-gram models show these preferences as well, though weaker. We do not find evidence that the models learned human word order preferences based on NP animacy in this case, but the experimental results suggest that the effects of animacy on this alternation might be more complex than previously believed.


\section{Dative alternation}

The \textsc{dative alternation} refers to the fact that in many cases the following forms are substitutable:
\ex. \label{ex:dative-alternation}
\a. The man gave the woman the book. [Double-object (\textsc{do}) construction] \label{ex:dative-alternation-do}
\b. The man gave the book to the woman. [Prepositional-object (\textsc{po}) construction]

The dative alternation is one of the most studied topics in syntax. The two constructions have been argued to convey subtly different meanings, with the \textsc{do} construction indicating caused possession and the \textsc{po} construction indicating caused motion \citep{green1974semantics,oehrle1976grammatical,gropen1989learnability,levin1993english}. However, the semantic preference of each construction appears to be only one factor among many when it comes to determining which form will be used in any given instance. Other factors include the animacy, definiteness, and length of the \textsc{theme} (\emph{the book} in Example~\ref{ex:dative-alternation}) and the \textsc{recipient} (\emph{the woman}) \citep{bresnan2007predicting}.

The human preferences in the dative alternation work out such that the NP which is more animate, definite, and short goes earlier. An extreme case is exemplified in~\ref{ex:dative-bad}: the sentences marked with $?$ are relatively dispreferred by native English speakers.
\ex. \label{ex:dative-bad}
\a. The man gave the woman a very old book that was about historical topics.
\b. $?$The man gave a very old book that was about historical topics to the woman.
\c. The man gave the book to a woman who was waiting patiently in the hallway.
\d. $?$The man gave a woman who was waiting patiently in the hallway a book.

In order to examine whether LSTMs show human-like preferences in the dative alternation, we designed 16 items on the pattern of \ref{ex:dative-alternation}, with 8 verbs of caused possession (such as \emph{give}) and 8 verbs of caused motion (such as \emph{throw}).\footnote{The preregistration for this experiment can be viewed at \url{https://aspredicted.org/ky9ne.pdf}.} In all items, the theme was inanimate and the recipient was animate. We manipulated the definiteness of the theme and the recipient using the articles \emph{the} and \emph{a}, and the length of the theme and the recipient by adding relative clauses to either or both. 

\begin{figure}
\centering
\includegraphics[scale=.45]{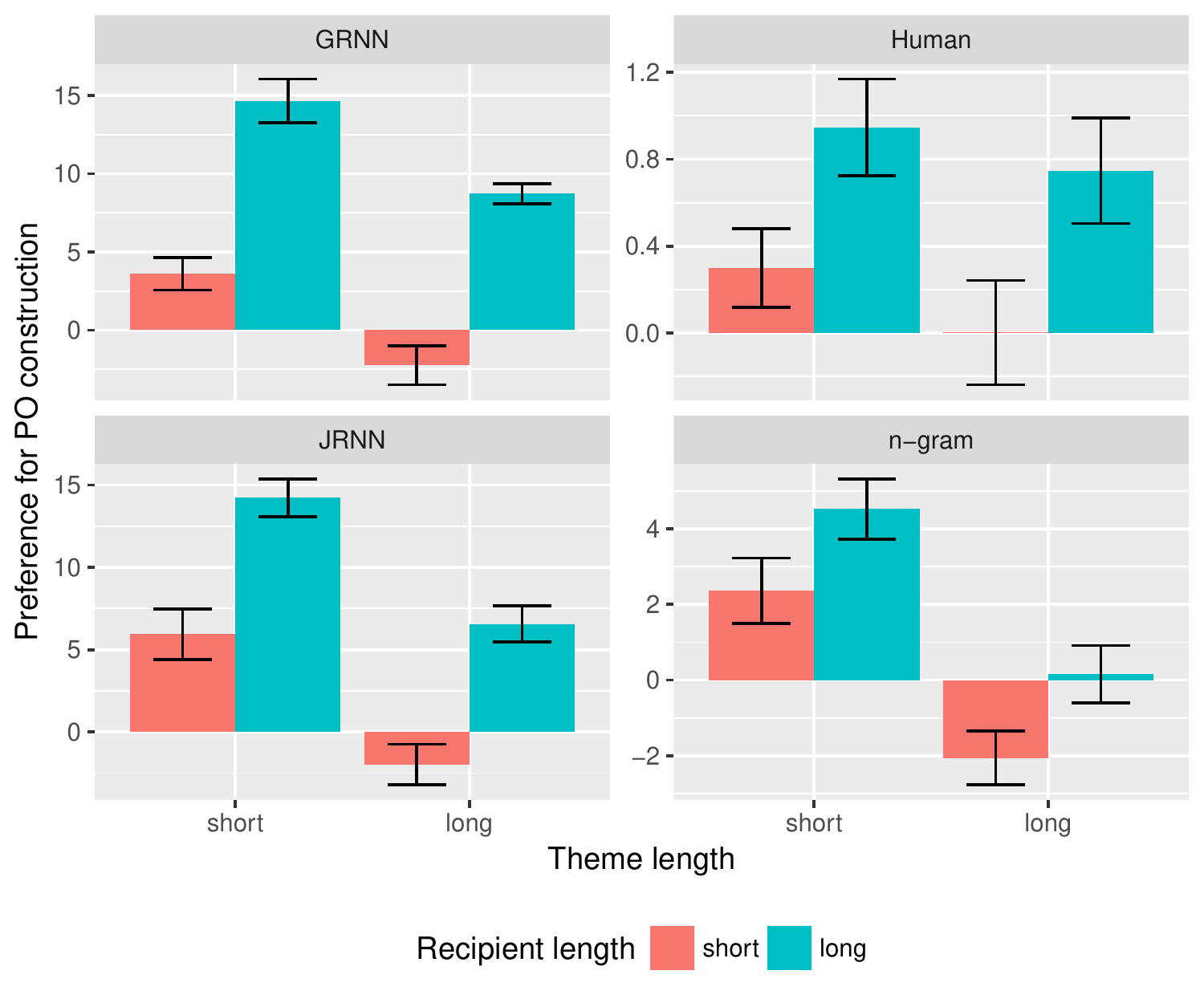}
\caption{Average PO preference by length of theme and recipient.}
\label{fig:dative-results-length}
\end{figure}

Figure~\ref{fig:dative-results-length} shows the strength of the models' preference for the \textsc{po} construction as a function of the length of the theme and recipient. The LSTMs have an overall preference for the PO form, which is mirrored in the human data. In addition, we see that a long recipient is strongly associated with a stronger preference for the PO form, and a long theme is strongly associated with a relative preference for the DO form, in line with human preferences. The $n$-gram baseline shows these effects but with a smaller magnitude, and without the overall PO preference shown by humans.

All interactions of length and word order are significant at $p<.001$ in all models, with the exception of the effect of recipient length in the $n$-gram model, where $p=0.01$. The effects of recipient and theme length are significantly weaker in the $n$-gram baseline than in \googlernn ($p<0.01$); for \gulordavarnn, the effect of recipient definiteness is significantly stronger than the $n$-gram baseline ($p=0.02$) but the effect of theme definiteness is not significantly stronger than in the $n$-gram baseline. In human data, the interaction of recipient length and word order is significant at $p<.001$ and the interaction of theme length and word order is significant at $p=.01$.

\begin{figure}
\centering
\includegraphics[scale=.45]{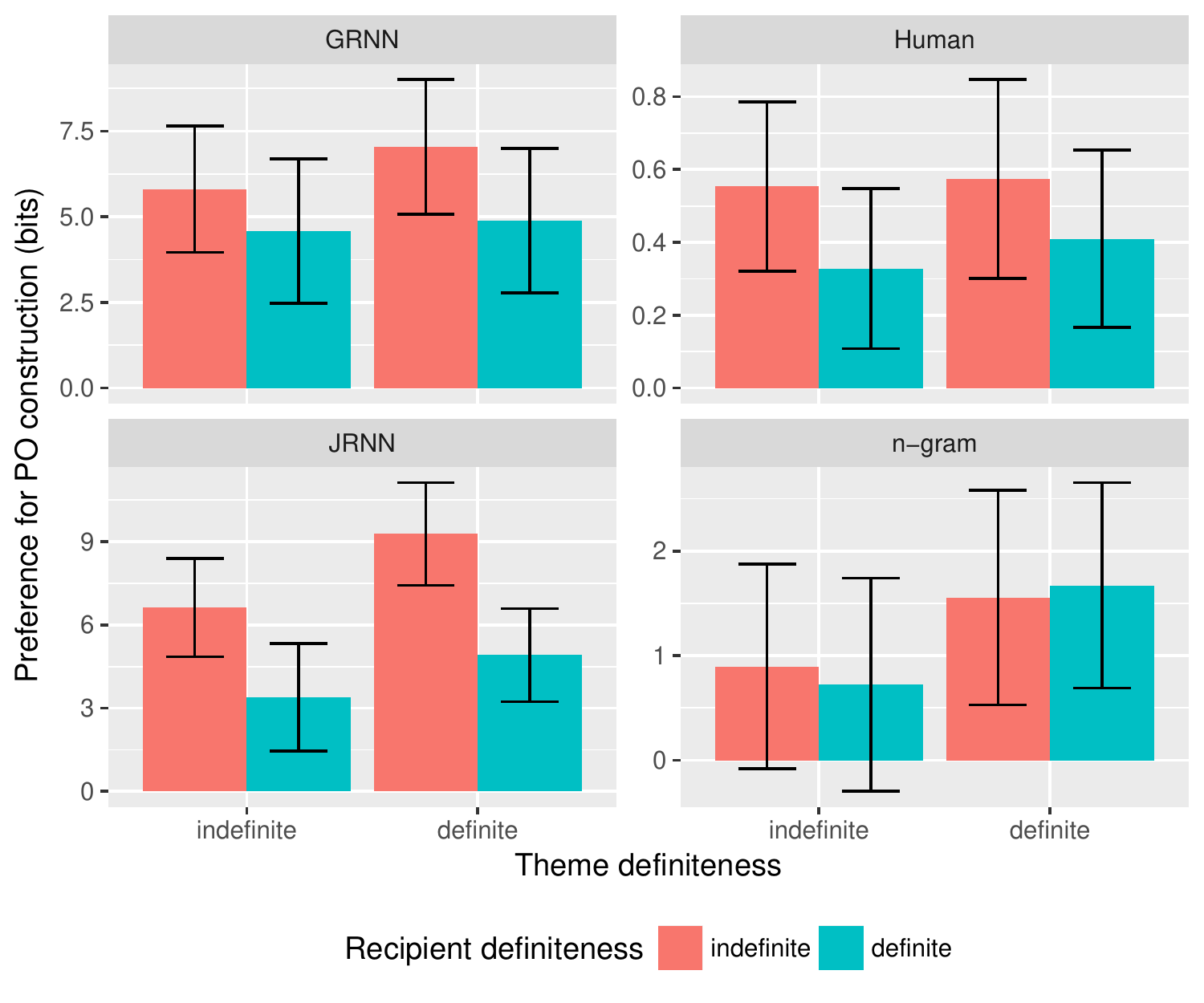}
\caption{Average PO preference by definiteness of theme and recipient.}
\label{fig:dative-results-definiteness}
\end{figure}

Now we turn to word order preferences based on NP definiteness. Figure~\ref{fig:dative-results-definiteness} shows the PO preference by the definiteness of the theme and recipient. In line with the linguistic literature, the PO preference is numerically smaller for definite recipients in both LSTM models and in human data, but not in the $n$-gram model. The interaction of recipient definiteness and word order is significant in the expected direction in \googlernn at $p<0.001$ and \gulordavarnn at $p=0.01$. Theme definiteness has a small positive interaction with word order (at $p<0.01$) for \googlernn, favoring the PO construction. These results are broadly in line with the linguistic literature, but they are not reflected in the human data for these experimental items: in the human ratings data, there are no significant interactions of definiteness and word order.

Overall, we find evidence for humanlike ordering preferences in the dative alternation with respect to length and definiteness of theme and recipient. The strongest effects which are most in line with the linguistic literature come from \googlernn. 

\section{Genitive alternation}

Similarly to the dative alternation, the \textsc{genitive alternation} involves two constructions with opposite word orders expressing similar meanings: 
\ex. \label{ex:genitive-alternation}
\a. The woman's house [\emph{s}-genitive, definite possessor] \label{ex:genitive-alternation-s}
\b. The house of the woman [\emph{of}-genitive, definite possessor] \label{ex:genitive-alternation-of}
\c. A woman's house [\emph{s}-genitive, indefinite possessor] 
\d. The house of a woman [\emph{of}-genitive, indefinite possessor] 

As in the dative alternation, whatever semantic difference exists between the two constructions is only one factor conditioning which form is used in each particular case. The other factors are the usual suspects: animacy, definiteness, and length of the \textsc{possessor} (\emph{the woman} in \ref{ex:genitive-alternation}) and \textsc{possessum} (\emph{the house} in \ref{ex:genitive-alternation}) \citep{kreyer2003genitive,rosenbach2003aspects,rosenbach2008animacy,shih2015rhythms}. 

In order to study the genitive alternation in RNNs, we designed 16 items on the pattern of \ref{ex:genitive-alternation}. We varied the definiteness and length of the possessor as in the dative alternation.\footnote{For both constructions to be legitimate syntactic options, the possesssum must be definite and unmodified by relative clauses.} We also varied the animacy of the possessor and possessum, between items.\footnote{The preregistration for this experiment can be viewed at \url{https://aspredicted.org/f2sk8.pdf}.}

\begin{figure}
\centering
\includegraphics[scale=.45]{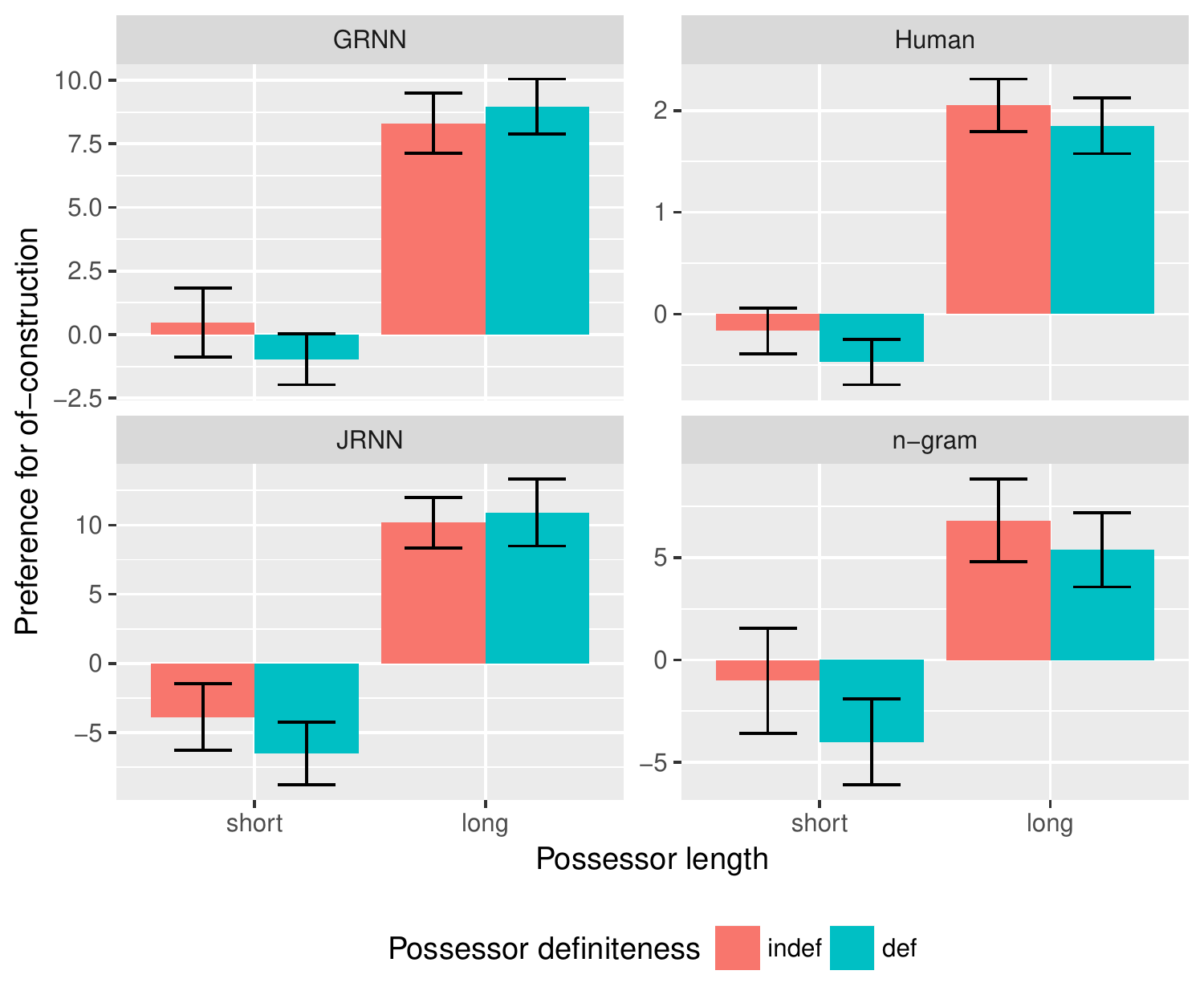}
\caption{Average \emph{of}-genitive preference by length and definiteness of possessor.}
\label{fig:genitive-etc}
\end{figure}

Figure~\ref{fig:genitive-etc} shows the RNNs' preferences for the \emph{of}-genitive form based on the definiteness and length of the possessor. In all models and in human data, we see that the \emph{of}-genitive is preferred generally when the possessor is long, and the \emph{s}-genitive when it is short. The interaction of possessor length with word order is significant in models and human data ($p<0.001$ in all cases). Turning to possessor definiteness, we see that it relatively favors the \emph{s}-genitive in human data and in the $n$-gram baseline, in line with the linguistic literature, but no such effect is found in the RNN models. However, the interaction of definiteness with word order is not significant in our data ($p=.09$ in human data and higher for the language models).

\begin{figure}
\centering
\includegraphics[scale=.45]{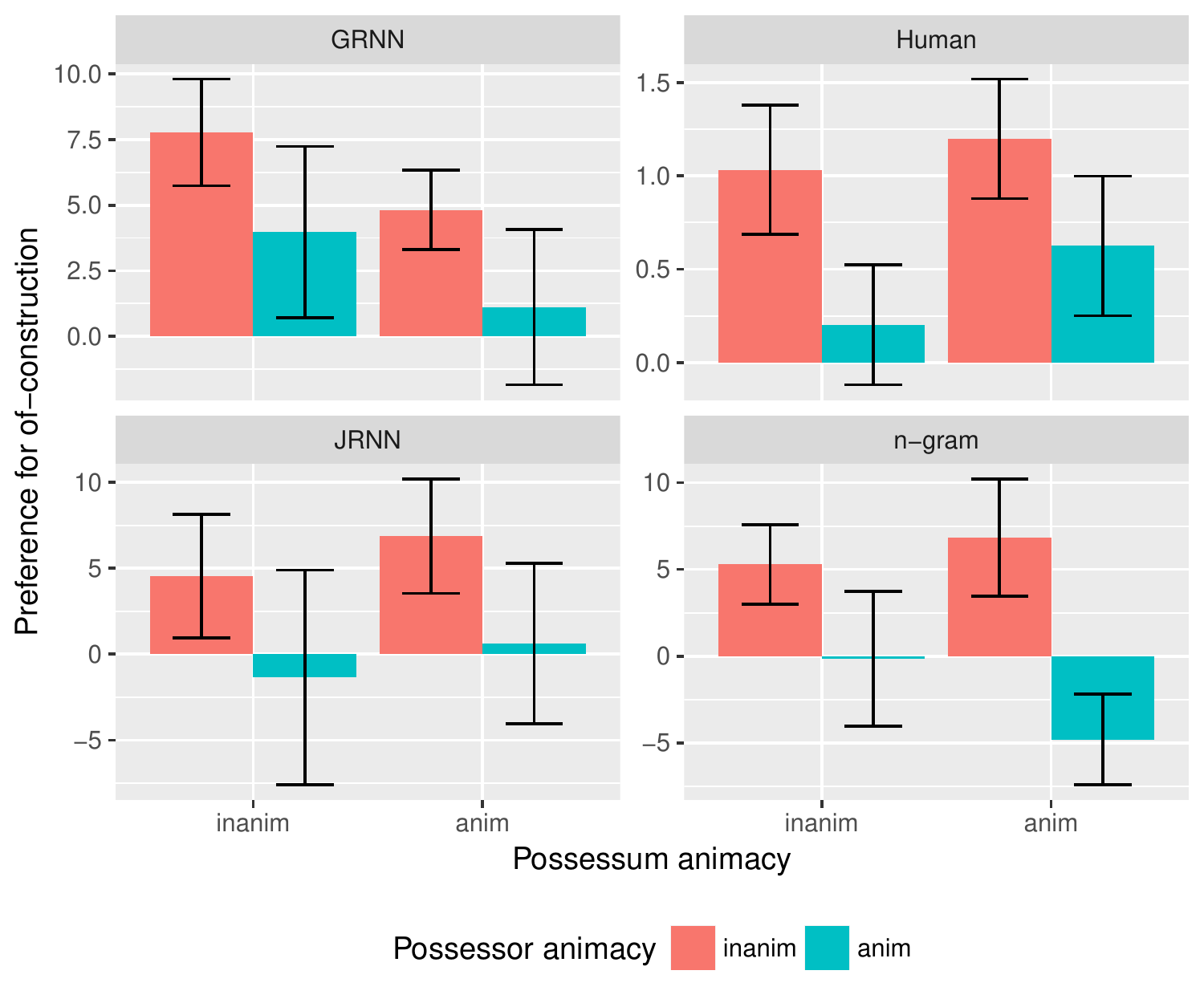}
\caption{Average \emph{of}-genitive preference by animacy of possessor and possessum.}
\label{fig:genitive-animacy}
\end{figure}

Now we turn to effects of animacy. Figure~\ref{fig:genitive-animacy} shows the \emph{of}-genitive preference by the animacy of the possessor and possessum. In all models, in line with human preferences, we see that possessor animacy favors the \emph{s}-genitive. The interaction of possessor animacy and word order is significant in \googlernn ($p=0.03$) and \gulordavarnn ($p<0.001$) but not in the $n$-gram baseline ($p=0.09$); the effect is significantly stronger in \googlernn than in the $n$-gram baseline ($p=0.02$) but not significantly stronger in \gulordavarnn. The effect of possessum animacy is more complex: it seems to favor the \emph{s}-genitive in \gulordavarnn, but the \emph{of}-genitive in the other models and in human preferences; in any case, the effect is small, and the interaction is not significant in any of the data collected here.

Overall, it appears that the LSTMs tested show humanlike order preferences in the genitive alternation when it comes to possessor length and animacy; they show more evidence for effects of possessor animacy than an $n$-gram baseline. They do not appear to pick up on definiteness preferences, but based on human experimental data these preferences might be weak in the first place.

\section{Discussion}

We have explored RNN language models' ability to represent soft word order preferences whose formulation requires abstract features such as animacy, definiteness, and length. We found that RNN language models generally do so, and outperform an $n$-gram baseline, indicating that they learn generalizations which are not trivially present in local co-occurrence statistics of words. The extent to which RNNs learn such preferences varies: effects of length are strongly and consistently represented, with weaker evidence for effects of animacy and definiteness. 

Much recent work has focused on whether RNNs can learn to represent discrete syntactic structures such as long-distance number agreement \citep{linzen2016assessing}, wh-dependencies \citep{mccoy2018revisiting,chowdhury2018rnn,wilcox2018what}, anaphora, negative polarity item licensing, and garden path sentences \citep{vanschijndel2018modeling,marvin2018targeted,futrell2018rnns}. The current work focuses on soft preferences which have been studied in quantitative syntax, and the abstract features that have been discovered to underly these preferences, finding that RNNs are able to represent many of the required features. The same features underlying these soft preferences in English often play a role in hard constraints in other languages \citep{bresnan2001soft}: thus our findings indicate that RNNs can learn crosslinguistically useful abstractions.

Our results also demonstrate that some of the key features underlying syntactic alternations can be learned from text data alone, without any particular innate bias toward such features. Qualifying this point, note that the language models we studied here were exposed to many more tokens of linguistic input than a typical child learner.

In addition to the results about RNNs, our work provides human data from directly controlled experimental manipulations of animacy, definiteness, and length in particle shift, genitive, and dative alternations. Our human acceptability ratings experiments have revealed some unexpected patterns, such as the sign reversal in the effect of animacy for long NPs in particle shift (Section~\ref{sec:particle-shift}), which should be investigated in more detail in future work.

An interesting question which has been studied in the functional linguistic literature is \emph{why} these particular word order preferences exist across languages. The preferences are often explained in terms of cognitive pressures on language comprehension and production. The short-before-long preference is most likely a manifestation of the pressure for short dependencies \citep{wasow2002postverbal}, which is motivated by working memory limitations in sentence processing \citep{gibson1998linguistic}; this word order preference is reversed in predominantly head-final languages such as Japanese, where there is a general preference for long constituents to come before short ones \citep{yamashita2001long}. The biases for animate and definite nouns to come early are usually linked to biases in the human language production process whereby words and constituents which are easier to produce come earlier \citep{bock1982toward}.

It is possible that these same cognitively motivated biases might also be present in RNNs. For example, \citet{futrell2017noisycontext} have argued that there should be a preference for short dependencies in any system that predicts words incrementally given lossy representations of the preceding context: since RNNs represent context using fixed-length vectors, their context representations must be lossy in this way. Furthermore, \citet{chang2009learning} has shown that the preference to place animate words earlier can arise in simple recurrent networks without this bias being present in training data, suggesting that RNNs may be subject to similar pressures to produce certain kinds of words earlier.

More generally, we have treated RNN language models essentially as human subjects delivering acceptability judgments, observing their behavior on carefully controlled linguistic stimuli rather than examining their internals. By using controlled experimental designs, we are able to control for factors such as sentence length and the particular lexical items in each sentence \citep[cf.][]{lau2017grammaticality}. We believe this approach will allow us to derive initial insight into the limits of what RNNs can do, and will guide work that explains the behavior we document here in terms of network internals.

\section*{Acknowledgments}
This work was supported in part by a gift from the NVIDIA corporation. RPL gratefully acknowledges support from the MIT-IBM AI Research Laboratory. All code and data is available at \url{https://github.com/langprocgroup/rnn_soft_constraints}.

\bibliographystyle{acl_natbib_nourl}
\bibliography{everything}

\end{document}